\documentclass{llncs}
\usepackage[ngerman,USenglish]{babel}
\usepackage[numbers]{natbib}
\pdfoutput=1

\makeatletter
\renewcommand\bibsection%
{
  \section*{\refname
    \@mkboth{\MakeUppercase{\refname}}{\MakeUppercase{\refname}}}
		\def\@biblabel##1{##1.}
			\small

}
\makeatother

\usepackage[T1]{fontenc}
\usepackage[latin1]{inputenc}
\usepackage[safe]{tipa}
\usepackage{lmodern}
\usepackage[hidelinks]{hyperref}

\begin{document}

\title{Are Style Guides Controlled Languages?}
\subtitle{The Case of Koenig \& Bauer AG}
\author{Karolina Suchowolec\thanks{I would like to thank Koenig \& Bauer AG for the permission to publish this paper, in particular Elmar Tober, who has been supervising the projects, and Sabine Lobach for providing up-to-date data. In addition, I would like to thank Klaus Schubert, Wolfgang Ziegler, and the anonymous reviewers for their critical comments.}}
\institute{University of Hildesheim, Germany \\
\email{karolina.suchowolec@uni-hildesheim.de}}

\maketitle
\begin{abstract}

Controlled languages for industrial application are often regarded as a response to the challenges of translation and multilingual communication \cite[pp. 52--53]{Crabbe2009}, \cite[p. 212]{Drewer2011}, \cite[pp. i--iii]{STE2013}. This paper presents a quite different approach taken by Koenig \& Bauer AG, where the main goal was the improvement of the authoring process for technical documentation. Most importantly, this paper explores the notion of a controlled language and demonstrates how style guides can emerge from non-linguistic considerations. Moreover, it shows the transition from loose language recommendations into precise and prescriptive rules and investigates whether such rules can be regarded as a full-fledged controlled language.

  \end{abstract}
\section{Introduction}
A considerable amount of research on controlled languages deals with English. In this paper, I examine an approach for an industrial application of a controlled language in German at the printing press manufacturer Koenig \& Bauer AG (KBA).\footnote{The author was employed at the company and implemented the terminology management as well as advised other projects on linguistic matters.} In comparison to well-known industrial examples such as CFE/CTE or ASD-STE100 \cite{STE2013}, \cite{Lockwood2000}, \cite[cf.][]{Crabbe2009}, KBA did not create an independent project on a controlled language. Instead, the company's main goal was to redefine the authoring process for technical documentation. Language standardization was needed to support this goal. I will explain this background in the first section. The following section will give a closer look at the results of different projects that originated different language rules.  These rules and style guides will illustrate how KBA unintentionally laid the foundations for a de facto controlled language. After showing some future directions for the project, I will discuss its results within the current theoretical frameworks.\par
Although this paper derives from a case study, its main goal is to explore the notion of a controlled language and to demonstrate how style guides can emerge from non-linguistic considerations. Moreover, this paper shows the transition from loose language recommendations to precise and prescriptive rules. It also investigates whether such rules can be regarded as a full-fledged controlled language.

\section{Background}
Koenig \& Bauer AG is one of the leading printing press manufacturers, operating from W\"urzburg, Germany, employing around 6,200 (2012) people worldwide. The variety of products covers a.\,o. web offset, sheetfed offset, and security printing. The annual reports\footnote{http://www.kba.com/investor-relations/berichte} indicate that over $80\,\%$ of the products are exported, which makes translation of documentation and localization of the press software an important step in the production process. Further, the web printing presses are unique custom-made production plants rather than standard models. In consequence, every operating manual is a unique document, describing specific features and the configuration of the press for a given customer. Such a document can be up to 800 pages long. It is translated into one language (national or regional language of the customer) and some customers require an additional English translation. In short it means: one press -- one manual in the source language (German) -- one manual in the target language. There are certainly some core parts of the press and some partial configurations recurring for many customers. In order to make the authoring process for technical documentation more efficient, it was, therefore, important to facilitate identifying, indexing, storing, retrieving, and combining the recurring information units. This task had to be accomplished before addressing any issues of translation. In other words: the company was mainly interested in increasing the reuse of the content.\par  
In order to redefine standard procedures for technical documentation, the company launched six projects in 2007. These projects contribute to the overarching goal of improving the reuse as follows \cite{Messaoudi2009}, \cite{Suchowolec2009}:
\begin{enumerate}
  \item CMS: Project aiming at the implementation of a content management system for editing, storing, retrieving and managing modularized content. \newline \mbox{TIM-RS\textsuperscript{\textregistered}} by Fischer Computertechnik\footnote{http://www.fct.de/de/loesungen/technische-dokumentation.html} was chosen as the CMS with the PI-Mod\footnote{http://www.pi-mod.de} as a data model for authoring. PI-Mod allows semantic as well as topic-based XML markup for text chunks and modular reusable information. This project provides the technology foundation for the reuse.
	\item Writers' manual: In this project, meta documentation of the CMS project was developed. The manual describes allowed XML elements: their content, dependencies, and, where necessary, their linguistic form. Moreover, it provides general recommendations on orthography and style for the technical documentation.
	\item Terminology management: A project on standardization of the specialized vocabulary. The goals were here to define standard procedures for the terminology management, to model the terminology database in SDL MultiTerm,\footnote{http://www.sdl.com/products/sdl-multiterm/desktop.html} and to develop linguistic rules for evaluation and selection of preferred terms.
	\item Translation: Most documents at KBA are written in German and the company does not have in-house translators; therefore the project defines general conditions for the translation services. In the context of this paper, the most important decisions are the prescribed use of Translation Memories, development of and adherence to the foreign-language equivalents to the provided German terminology as well as development of and adherence to general style guides for each foreign language as needed.
	\item Graphics: Due to the variability of the products, a full reuse of graphics in operating manuals is impossible. This project should nonetheless define common standards regarding the exporting of graphics from CAD systems, further processing and managing in the CMS environment.
	\item Parts catalog: In contrast to the above, this project focuses not on reuse, but on making the editing and publishing of the parts catalog more simple and transparent.
\end{enumerate}
As we see, there was no particular project for developing a controlled language as such. Yet, relevant issues and requirements concerning language use are scattered across all projects.  

\section{Language Constraints at KBA}

KBA chose a modular approach towards development of technical documentation, which means that the documents are built up by combining different text modules and chunks. These chunks can already be stored in a database; however, they could be written by different writers at a different time and for different products. The reuse principle requires not only homogeneous layout and typography, but also explicit linguistic rules for all writers involved in order to keep or achieve consistent language. Not only does consistent language make the manuals more readable for the target (human) audience, but at the same time it also supports the implementation of the reuse technology; for instance: the vocabulary is used to index modules and chunks. Therefore, consistent terminology improves the (machine-aided) retrieval.\par
In conclusion, it was the variability of the products as well as the desire to make the process of the technical writing more efficient that led to the development of the style/syntax and lexical rules. In addition, improving the readability and translatability was an important aspect of standardizing the vocabulary.\par
In the following section, I will elaborate on the linguistic results of the projects 1--4 described above dividing them into issues concerning lexicon and syntax/style. By giving some examples of the rules and by describing the rule-formation process, I would like to draw attention to some aspects that will be subsequently crucial to the understanding of a controlled language. First, KBA went beyond the regular terminology management by developing a rule-based selection of preferred terms. Second, some imprecise style guides emerged from non-linguistic considerations. Thanks to human and later machine-aided editing, the rules have become more explicit and prescriptive. The following sections will illustrate this transition.   

  \subsection{Lexical Level}
KBA imposed two constraints to the standardization of the specialized vocabulary.\footnote{This section is based on \cite{Suchowolec2009}.} The first constraint limited the in-house standardization to the German terms only. The second one restricted the scope of standardization to the specific printing vocabulary, leaving out the more general technical terms or terms used in other domains such as economics. Importantly, KBA required the selection of preferred terms to be reproducible and transparent to the writers. Therefore, linguistic criteria were developed for evaluating existing terms, which can be applied for coining new terms as well. Using a corpus of company's vocabulary, semantic and morphological aspects of the word formation patterns were identified and evaluated. An essential part was to identify the patterns leading to ambiguity and synonymy, for instance:
\begin{itemize}
  \item Ambiguity: Nominalization using the suffix \emph{-ung} can both indicate a process or a device performing the process: \emph{wenden} (\emph{to turn a sheet for perfecting}) $\rightarrow$ \emph{Wendung} (\emph{the process of perfecting or a perfecting unit}).
	\item Synonymy:
	  \begin{itemize}
		  \item A process can be expressed through nominalization with suffix \emph{-ung} (see above) or conversion of the infinite verb form: \emph{wenden} $\rightarrow$ \emph{Wend{\bfseries ung}} vs. \emph{Wend{\bfseries en}}. 
			\item Different features of a concept can be stressed in synonymous terms: \emph{{\bfseries Chrom}walze} (\emph{chrome roller}, focus on material) vs. \emph{{\bfseries Feuchtreib}walze} (\emph{dampener distributor roller}, focus on function).
			\item The hypernymy can be explicitly stated using the hypernym in compound nouns, or this relation can be stated implicitly: \emph{Farbreib{\bfseries walze}} (\emph{oscillating ink roller}) vs. \emph{Farbreib{\bfseries er}} (*\emph{ink oscillator}).
		\end{itemize}
\end{itemize}

One of the goals in developing the linguistic criteria/rules was to achieve a one-to-one relation between the word formation pattern and the semantic class of the objects. In other words, the \emph{signifiant} should indicate the class of the \emph{signifi\'e}, as illustrated by the following rules:
\begin{itemize}
  \item Use conversion to indicate the process: \emph{Wenden}.
	\item Use the nominalization with \emph{-ung} or \emph{-or/-er} to indicate a (complex) device: \emph{Wendung}, \emph{L\"angsschneider} (\emph{slitter}).\footnote{The corpus indicates a complementary distribution of both patterns with only few exceptions.} 
	\item Use hypernym for composition of (less complex) parts: \emph{Farbreibwalze}, \emph{Schneid\-messer} (\emph{cutting knife}).
	\item Use the following ranking of features to be included in the term: 1) function 2) object 3) working principle 4) shape 5) material 6) temporal, graduate, internal features \cite[cf.][p. 14]{Reinhardt1992}.
	\item Do not use more than 4 lexical morphemes in a compound noun, 3 morphemes are preferred.
\end{itemize}  

These criteria are intended to be an assistance rather than absolute rules for selecting the preferred terms. In case they lead to the selection of extremely unusual forms, established terms are preferred.\par
These rules, steps and roles in the terminology management process as well as the definition of the data structure for the terminology database were fixed in a terminology manual. Only after this definition task was completed, the actual standardization of the lexicon began. Contrary to the manual, which recommended management similar to  model B of ISO 15188 \cite{ISO15188}, the initial standardization followed rather model D. The first terminology draft was proposed by the terminologist and consulted with the technical writers. The final form was released only after consulting with the constructing engineers, who gave their feedback on every term in the draft.\par
Applying the theoretical principles of terminology management by W\"uster \cite{Wuester1991}, \cite[cf.][]{Picht2014}, the standardization process was concept-driven. Manually extracted terms were first arranged into concept systems (multiple arrangement of one term was allowed) and then given a definition. Only after both the concept systems and the definition were specified, the preferred term was chosen according to the rules described above.\par
SDL MultiTerm was used from the very beginning. In addition, the initial management employed MS PowerPoint (concept systems) and MS Excel (definitions and synonyms), in order to facilitate the feedback by the engineers. Moreover, the workflow software quickTerm by Kaleidoscope\footnote{http://www.kaleidoscope.at/Deutsch/Software/QuickTerm/quickterm.php} is now being implemented, which will help to shift back to the originally intended management as in the model B. \par
Currently, the database contains 614 concepts (1689 terms), covering almost entirely the printing specific vocabulary for the operating manuals. Roughly $30\,\%$ of the German terms have an approved foreign-language equivalent in one or more of the following languages: English, French, Dutch, Russian, Swedish, and Spanish. The equivalents are provided by the translation services based on the given German terminology and are not double-checked by KBA before final release. However, quickTerm allows users to send their feedback on all languages.          
	
	\subsection{Syntactic and Stylistic Level}
Unlike the lexical level, where an effort was put in the linguistic evaluation of the corpus data and development of systematic rules for word formation, the development of the stylistic and syntactic rules was a byproduct of the implementation of PI-Mod.\par
PI-Mod uses XML elements to mark up information in a text according to its semantics, for instance as <step>, <descriptive>, <precondition>, <cause>, or <solution>. After agreeing on the elements needed for operating manuals, the standardization of syntax and style for some elements began. This standardization was necessary for a similar reason as the lexical one: competing syntactic patterns were in use, for instance, imperative verb form vs. infinite verb form used as imperative in <step> (in <action>), full sentence vs. ellipsis in <cause> etc.\par
There is no record of the decision-making process for these rules, as the focus of the CMS project was on the technical rather than linguistic specification \cite{Messaoudi2009}. However, the original writers' manual and personal communication indicate that the prescribed style patterns were a combination of so-called good practice for technical documentation, implicit or explicit but merely oral arrangements among the writers, and the standard examples used in the general PI-Mod specification.\par
The rules can be divided into general and element specific rules. Here are some examples of the original recommendations developed during the CMS project:
\begin{itemize}
  \item Avoid Passive Voice.
	\item Structure information logically, for instance if -- then, or condition -- step.
	\item Element <step> (as child element of <action>):\par
	Use the formal imperative verb form (`Sie').
	\item Element <symptom>:\par
	Write from user's perspective, do not use questions.\par
	Example: \emph{Mastarm f\"ahrt nicht richtig hoch.} [sentence with a finite verb]\footnote{Information in brackets was not indicated in the manual.} 
	\item Element <cause> (as child element of <safetyadvice>):\par 
	Name the cause of the hazard with one word or in a short and expressive sentence. Use an exclamation point.\par
	Example 1: \emph{Verbr\"uhungsgefahr durch herausspritzendes \"Ol!} [ellipsis, no finite verb]\par
	Example 2: \emph{\"Oldruck in Arbeitshydraulik kann Manometer zerst\"oren!} [sentence with a finite verb]
	\item Element <cause> (as child element of <errordescription>):\par
	Describe the cause of an error in one word or in a short and expressive sentence.\par
	Example 1: \emph{Kein Kraftstoff im Tank.} [ellipsis, no finite verb]\par 
	Example 2: \emph{Pumpe hat zu wenig Leistung.} [sentence with a finite verb]
\end{itemize} 

Initially, the syntactic and stylistic rules were enforced by human editing within the department. Already at that point it became clear that some rules remained ambiguous. As demonstrated in the examples above, some rules were linguistically imprecise, some lacked an explicit form, some depended on (sometimes contradicting) examples and did not indicate whether they were recommendations or prescriptions. The room for interpretation lowered the consistency of the texts and in consequence -- the reuse. Therefore, the rules and examples in the writers' manual have been gradually replaced by more precise ones, for instance:
  \begin{itemize}
	  \item Element <cause> (as child element of <safetyadvice>):\par
		Use ellipsis (construction with no finite verb). Do not use full sentences with verbs. Name the cause of the hazard with one word (\emph{Verbr\"uhrungsgefahr, Verbrennungsgefahr}). Use an exclamation point. \par
		Positive example: \emph{Maschinenschaden durch liegengebliebenes Werkzeug!}\par
		Negative example: \emph{Liegengebliebenes Werkzeug f\"uhrt zu Maschinenschaden.}\par
	  \item Element <cause> (as child element of <errordescription>):\par
		Use a full sentence with a verb. Do not use ellipsis. Use a period at the end of the sentence.\par
		Positive example: \emph{Kein Kraftstoff ist im Tank.}\par 
		Negative example: \emph{Kein Kraftstoff im Tank.}
	\end{itemize}
	
Further specification of the rules has been reinforced by the implementation of a controlled language checker (CLC, Acrolinx\footnote{http://www.acrolinx.com}), since the rules had to be easily transformed into a machine-readable form. Hence, the syntactic/stylistic level is being further consolidated.\par
Although not used from the beginning, a CLC was considered a medium-term goal. At this time, the system is being implemented to fit into the already existing linguistic environment. With respect to style and syntax, KBA rules are being mapped to the standard Acrolinx set of rules and the system is being checked in a test environment. Practical application is expected not earlier than in the summer 2014. Despite offering solutions for terminology management, Acrolinx will solely be used for proofreading. SDL MultiTerm will remain the primary source for the lexical level of the language. 

\section{Prospects}

The four crucial projects resulting in the linguistic rules -- CMS, terminology, writers' manual and translation -- are completed. Yet, the maintenance and further development of the rules is an on-going process. Other tasks like the productive use of a CLC are still to be accomplished. More importantly, described development is normative only to the department of technical documentation; however other departments can obviously benefit from the use of language restrictions as well. Some applications might regard CAD models, ERM (mainly terminology), and press software (both terminology and syntax/style). Although some ways of implementing a controlled language to these applications have already been explored, they still remain a challenge for the future.\par
Overall the main goal of modular approach and reuse still has to be evaluated. KBA is going to track the reuse applying the REx method \cite{Oberle2012}. Resulting data could then be used for further interpretation of the linguistic contribution to the (improved) reuse.\footnote{Wolfgang Ziegler, email communication (February 20, 2014)}   

\section{Discussion}

Before any specific issues concerning language at KBA can be discussed, it needs to be determined whether the rules and developments described above constitute a controlled language.\par
The term \emph{controlled language} is not used in any internal documents to describe the lexical or syntactic rules for technical documentation. Rather, there is an issue of terminology and writing rules (\emph{Schreibregeln}) -- both being treated separately. The constraints on the language are put into an overarching context of standardization, just like XML markup, modular editing or data indexing in the CMS. This might be the result of the distribution of the linguistic decisions over the four projects. In consequence, an explicit notion of a \emph{controlled language} has not yet emerged.\par
From the theoretical point of view, however, the linguistic constraints at KBA satisfy all of the criteria by Kuhn \cite[p. 123]{Kuhn2014}: They are based on just one language and restrict its system on the orthographic, morphological, lexical, syntactic and textual level, combining both prescriptive and proscriptive rules \cite[p. 228]{Drewer2011}. Although no empirical studies are available, we can assume that the output is easily recognizable as German to an expert familiar with the sublanguage of printing. And finally, although the already established forms that had developed in an unsystematic way were preferred, the codification of the forms in manuals was a deliberate and, to a certain extent, systematic process. Therefore, we can regard these linguistic constraints as a controlled language. \par
Having clarified the status of the linguistic constraints at KBA as controlled language, we can try to determine its type.\par
We can apply to it the categories human-oriented and machine-oriented as proposed by Huijsen \cite{Huijsen1998a}. The development of the controlled language at KBA indicates an expansion of originally predominant human-oriented language to a language that comprises both human- and machine-relevant features, but a clear determination is  difficult. Both categories seem to be tendencies rather than a dichotomy, which supports the main view in the literature \cite[p. 2]{Huijsen1998a}, \cite[cf.][p. 125]{Kuhn2014}.\par
Moreover, we can examine further motives for this controlled language. There are three main groups of motives discussed in the literature: 1) to improve the communication among humans, 2) to improve the translation, and 3) to represent formal notations \cite[p. 125]{Kuhn2014}, \cite[cf.][p. 1]{Huijsen1998a}, \cite[cf.][p. 225]{Schubert2001b}, \cite[cf.][p. 134]{Schubert2009}, leading Kuhn to postulate three main types of controlled languages \cite[p. 125]{Kuhn2014}. As demonstrated above, these motives were not of primary concern for the style rules at KBA, where the increase of reuse was the main goal. The reuse issue of a controlled language is not a new topic in the literature. It has been discussed in several case studies (CTE: \cite[p. 422]{Gallup1993}, \cite[p. 194]{Lockwood2000}). It is also mentioned in some general overviews on controlled languages, but it rarely seems to be the major aspect discussed (in detail: \cite[pp. 206--207]{Drewer2011}, briefly:  \cite[p. 11]{Hebling2002}, \cite[p. 248]{Nyberg2003}, \cite[p. 62]{Ramirez2012}, left out: \cite[pp. i--iii]{STE2013}, \cite{Crabbe2009}). Clearly, the reuse motive is missing in the classification by Kuhn. This might suggest the fourth type of a controlled language -- to improve (the efficiency of) the authoring process for technical documentation -- an issue which seems to be underrepresented in the literature.\par
Certainly, a legitimate question may arise whether a controlled language for the content reuse serves solely the purpose of ensuring the readability of the text for the target audience. As such, it could be subsumed under the first type of controlled languages. As we saw, controlled language at KBA, in particular the vocabulary, ensures not only the consistency of a text, but also the retrieval of modules, and, hence, contributes to the reuse in various ways. It is not to say that KBA was unaware that the improved clarity and precision of a controlled language have impact on the readability for the target audience, as well as improve (human and machine-aided) translatability and help control company's liability in case of damage to persons or facilities \cite{Messaoudi2009}, \cite{Suchowolec2009}, \cite[cf.][p. 443]{Kittredge2003}, \cite[cf.][pp. 225--230]{Schubert2001b}. These secondary motives were certainly important for the support of the projects at different management levels in the company \cite[cf.][Modul 1]{DTT2014}. Yet these general considerations only became tangible once the modular editing for improving the reuse started to be implemented, since, in an obvious way, the modular approach jeopardizes the consistency with all the consequences.\par 
What is more, the aspect of human understanding can also be found in controlled languages for translation -- a controlled language ensures a.\,o. a correct and readable output of (human, machine-aided or machine) translation. Nonetheless, controlled languages for translation constitute their own type. Altogether, taken the classification by Kuhn as a starting point, it seems, by analogy to the controlled languages for translation, reasonable to add the fourth type of controlled languages mentioned above to the classification.\par    
\par 
Another matter to discuss is the use of CLC and a general acceptance of a controlled language. Based on observation, the need for the application of language rules is generally understood and accepted by the technical writers. First tests of CLC seem to conform this. However, ambiguous terms lower the precision of the CLC, which is consistent with the difficulties mentioned in the literature \cite[p. 8]{Huijsen1998a}, \cite[p. 252]{Nyberg2003}. Splitting the vocabulary into domains, as suggested by the software provider and by some authors \cite[p. 276]{Nyberg2003} seems to lack of a good splitting criterion that would not cause too much term overlap between the domains, requiring some essential changes in the database definition and a reconsideration of the concept-driven terminology management. Currently, different models for software feedback and score in case of ambiguity are being tested. Overall, the problem remains unsolved. The main question concerning this type of feedback is whether it will have an impact on the effectiveness and on general acceptance of a CLC by the writers.\par
Looking from a broader perspective, the case of KBA sheds some light on the more general issue of the acceptance of a controlled language by the writers. Godden reports on the difficulties with the editor-centered model for CASL at General Motors, where the writers not previously trained in the rules were reluctant to accepting the changes proposed by the editor. Some better results were achieved by a hybrid model, requiring prior training of the writers \cite{Godden2000}, \cite[cf.][pp. 215--216]{Drewer2011}, \cite[cf.][pp. 107--112]{Hebling2002}. We do not have similar data concerning machine editing at KBA yet, but the experience with the human editing shows a promising degree of acceptance of the language rules and proposed changes in case of a clear violation of the rules. I presume that there is a link between cooperative approach to the definition of the rules, resulting on the one hand in a `sense of community' and on the other in a better comprehension of the rules, leading altogether to an increase of acceptance. There has already been an awareness of this phenomenon, particularly in the literature on terminology management \cite[p. 166]{Drewer2011}, \cite[cf.][p. 63]{Ramirez2012} but empirical studies are still needed.   
		
\section{Concluding Remarks}

The case of KBA shows that the notion of a controlled language needs to be reexamined. It can emerge as a result of deliberate decisions that are not necessarily labeled and conceived as a design of a controlled language. The distinction between controlled languages and style guides is indeed vague \cite[p. 124]{Kuhn2014} and needs to be examined on a case by case basis, and perhaps only at a given point in time. This also suggests that there might be more companies having language regulations that do not see themselves as being concerned by the discourse of controlled languages. Furthermore, a new type of a controlled language -- to improve the authoring process for technical documentation -- needs to be considered.\par
From the practical point of view, the case of KBA shows that the development in technology has made a controlled language more attainable. There is no longer a need for custom-built software solutions to implement the specifications of a controlled language. There is an array of standard software that can be customized to fit specific needs, which makes a controlled language easier and more affordable to develop and implement, opening the field to medium-size companies.

\bibliography{cnl}

\end{document}